\documentclass{article}
\usepackage{ijcai20}

\usepackage{times}
\usepackage{soul}
\usepackage{url}
\usepackage[utf8]{inputenc}
\usepackage[small]{caption}
\usepackage{graphicx}
\usepackage{amsmath}
\usepackage{amsthm}
\usepackage{booktabs}
\usepackage{algorithm}
\usepackage{algorithmic}
\usepackage{makecell}
\usepackage{multirow}
\title{A Graph-based Interactive Reasoning for Human-Object Interaction Detection}

\author{
Dongming Yang$^1$\And
Yuexian Zou$^{1,2}$\footnote{Contact Author}\\
\affiliations
$^1$School of ECE, Peking University, Shenzhen, China\\
$^2$Peng Cheng Laboratory, Shenzhen, China\\
\emails
\{yangdongming, zouyx\}@pku.edu.cn
}
\begin{document}

\maketitle

\begin{abstract}
  Human-Object Interaction (HOI) detection devotes to learn how humans interact with surrounding objects via inferring triplets of $\langle$ human, verb, object $\rangle$. However, recent HOI detection methods mostly rely on additional annotations (e.g., human pose) and neglect powerful interactive reasoning beyond convolutions. In this paper, we present a novel graph-based interactive reasoning model called Interactive Graph (abbr. in-Graph) to infer HOIs, in which interactive semantics implied among visual targets are efficiently exploited. The proposed model consists of a project function that maps related targets from convolution space to a graph-based semantic space, a message passing process propagating semantics among all nodes and an update function transforming the reasoned nodes back to convolution space. Furthermore, we construct a new framework to assemble in-Graph models for detecting HOIs, namely in-GraphNet. Beyond inferring HOIs using instance features respectively, the framework dynamically parses pairwise interactive semantics among visual targets by integrating two-level in-Graphs, i.e., scene-wide and instance-wide in-Graphs. Our framework is end-to-end trainable and free from costly annotations like human pose. Extensive experiments show that our proposed framework outperforms existing HOI detection methods on both V-COCO and HICO-DET benchmarks and improves the baseline about 9.4\% and 15\% relatively, validating its efficacy in detecting HOIs.
\end{abstract}

\section{Introduction}

The task of Human-Object Interaction (HOI) detection aims to localize and classify triplets of $\langle$ human, verb, object $\rangle$ from a still image. Beyond detecting and comprehending instances, e.g., object detection \cite{ren2017faster,he2017mask}, segmentation \cite{yang2018denseaspp} and human pose estimation \cite{fang2017rmpe}, detecting HOIs requires a deeper understanding of visual semantics to depict complex relationships between human-object pairs. HOI detection is related to action recognition \cite{sharma2015action} but presents different challenges, e.g., an individual can simultaneously take multiple interactions with surrounding objects. Besides, associating ever-changing roles with various objects leads to finer-grained and diverse samples of interactions.

\begin{figure}
	\centering
		\includegraphics[scale=.28]{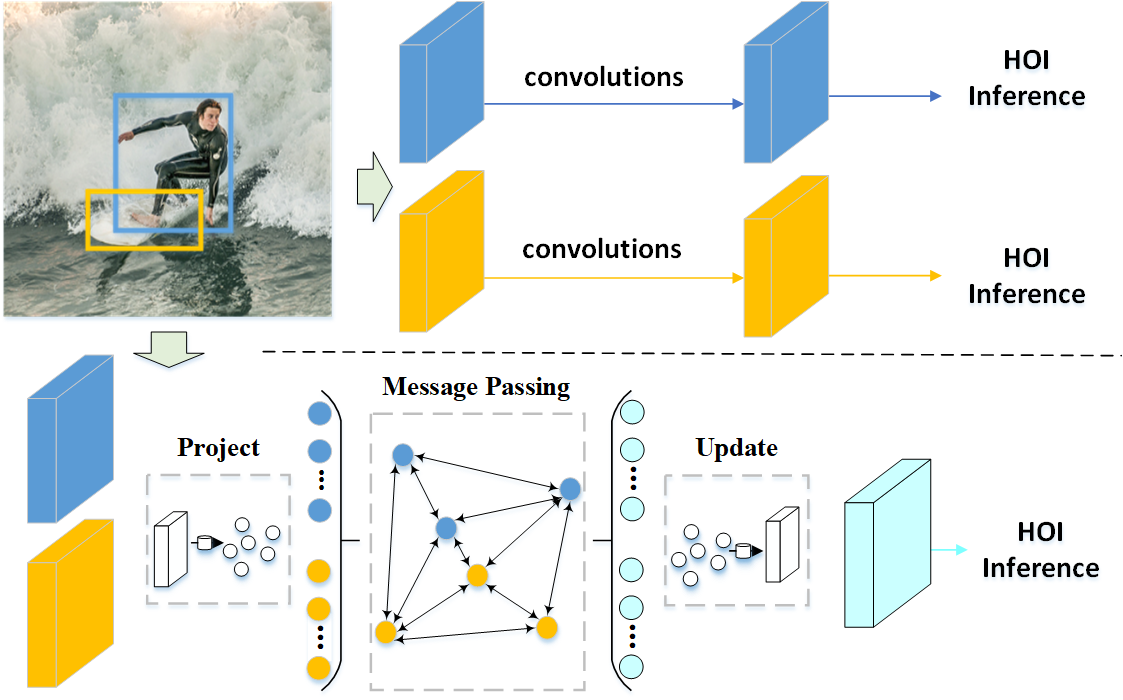}
	\caption{A simplified schematic of how our framework facilitates HOI detection. The baseline framework (above) adopts convolutional features directly to infer HOIs and neglects interactive reasoning. We present a novel model (bottom) that reasons interactive semantics among visual targets to learn HOI-specific representation.}
	\label{FIG:1}
\end{figure}

Most existing HOI detection approaches infer HOIs by directly employing the appearance features of a person and an object extracted from Convolutional Neural Networks (CNNs) \cite{he2016deep}, which may neglect the detailed and high-level interactive semantics implied between the related targets. To remedy the limitation above, a number of algorithms exploit additional contextual cues from the image such as human intention \cite{xu2018interact} and attention boxes \cite{kolesnikov2018detecting}. More recently, several works have taken advantage of additional annotations, e.g., human pose \cite{xu2018interact,li2019transferable} and body parts \cite{zhou2019relation,wan2019pose}. Although incorporating contextual cues and annotations generally benefits feature expression, it brings several drawbacks.

Firstly, stack of convolutions and contextual cues are deficient in modelling HOIs since recognizing HOIs requires reasoning beyond feature extraction. However, dominant methods are limited by treating each visual component separately without considering crucial semantic dependencies among related targets (i.e., scene, human and object). Secondly, employing additional human pose and body parts in HOI detection algorithms brings a large computational burden.

To address these issues, we propose a novel graph-based model called Interactive Graph (abbr. in-Graph) to infer HOIs by reasoning and integrating strong interactive semantics among scene, human and object. As illustrated in Figure~\ref{FIG:1}, our model goes beyond current approaches lacking the capability to reason interactive semantics. In particular, in-Graph model contains three core procedures, i.e., a project function, a message passing process and an update function. Here, the project function generates a unified space to make two related targets syncretic and interoperable. The message passing process further integrates semantic information by propagating messages among nodes. Finally, the update function transforms the reasoned nodes to convolution space, providing enhanced representation for HOI-specific modeling.

Based on the proposed in-Graph model, we then offer a general framework referred to as in-GraphNet to implicitly parse scene-wide interactive semantics and instance-wide interactive semantics for inferring HOIs rather than treat each visual target separately. Concretely, the proposed in-GraphNet is a multi-stream network assembling two-level in-Graph models (i.e., scene-wide in-Graph and instance-wide in-Graph). The final HOI predictions are made by combining all exploited semantics. Moreover, our framework is free from additional annotations such as human pose.

We perform extensive experiments on two public benchmarks, i.e., V-COCO  \cite{yatskar2016situation} dataset and HICO-DET \cite{chao2018learning} dataset. Our method provides obvious performance gain compared with the baseline and outperforms the state-of-the-art methods (both pose-free and pose-based methods) by a sizable margin. We also provide detailed ablation studies of our method to facilitate the future research.

\section{Related Work}
\subsection{Contextual Cues in HOI Detection}

The early human activity recognition \cite{idrees2017thumos} task is confined to scenes containing single human-centric action and ignores spatial localization of the person and related object. Therefore, Gupta \cite{yatskar2016situation} introduced visual semantic role labeling to learn interactions between human and object. HO-RCNN \cite{chao2018learning} introduced a three-branch architecture with one branch each for a human candidate, an object candidate, and an interaction pattern encoding the spatial position of the human and object. Recently, several works have taken advantage of contextual cues and detailed annotations to improve HOI detection. Auxiliary boxes \cite{gkioxari2015contextual} were employed to encode context regions from the human bounding boxes. InteractNet \cite{gkioxari2018detecting} extended the object detector Faster R-CNN \cite{ren2017faster} with an additional branch and estimated an action-specific density map to identify the locations of interacted objects. iHOI \cite{xu2018interact} utilized human gaze to guide the attended contextual regions in a weakly-supervised setting for learning HOIs.

\begin{figure*}
	\centering
		\includegraphics[scale=.25]{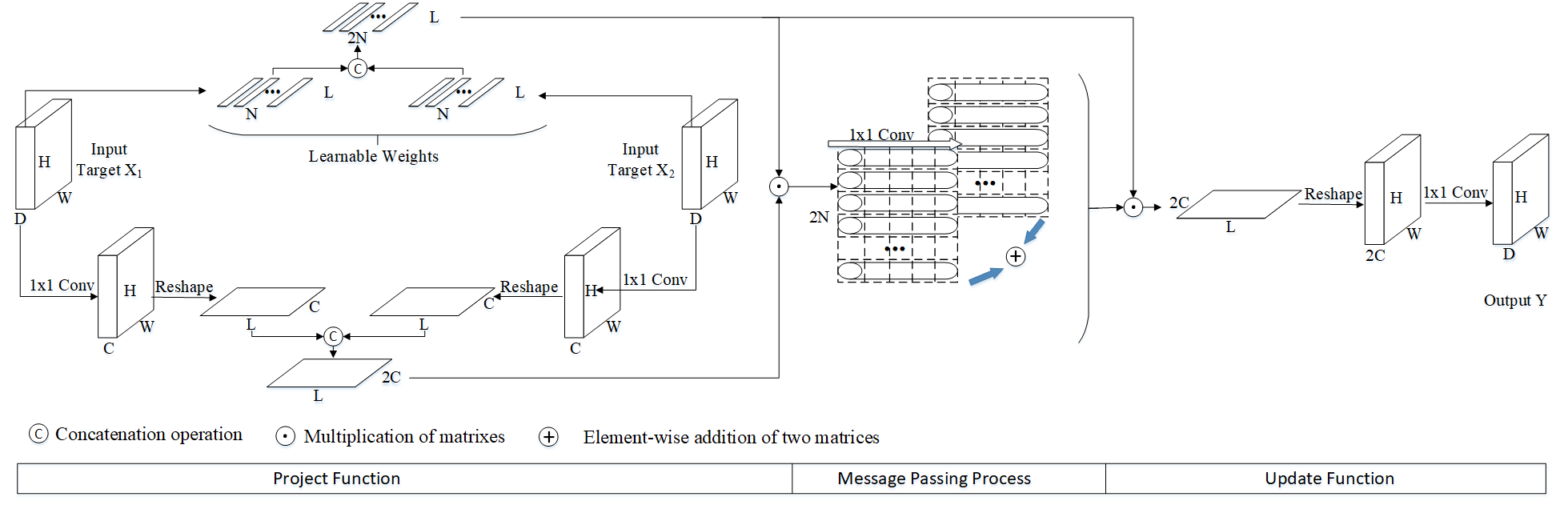}
	\caption{The detailed design of proposed in-Graph model. The model consists of three core procedures, which are a project function, a message passing process and an update function. The in-Graph takes convolutional features of two visual targets as inputs and reasons strong interactive semantics between them by a graph-based structure. Finally, reasoned semantics are output in same form as inputs.}
	\label{FIG:2}
\end{figure*}

%\subsubsection{Human pose}
%\paragraph{Human pose.}
Very recently, human pose \cite{fang2017rmpe} has been widely adopted as an additional cue to tackle HOI detection. Pair-wise human parts correlation \cite{fang2018pairwise} was exploited to learn HOIs. TIN \cite{li2019transferable} combined human pose and spatial configuration to encode pose configuration maps. PMFNet \cite{wan2019pose} developed a multi-branch network to learn a pose-augmented relation representation to incorporate interaction context, object features and detailed human parts. RPNN \cite{zhou2019relation} introduced an object-bodypart graph and a human-bodypart graph to capture relationships between body parts and surrounding instances (i.e., human and object).

Although extracting contextual evidence benefits feature expression, it is not favored since additional annotations and computation are indispensable. For example, pose-based approaches are inseparable from pre-trained human pose estimators like Mask R-CNN \cite{he2017mask} and AlphaPose \cite{fang2017rmpe}, which bring large workload and computational burden.

\subsection{Semantic Reasoning}

%\subsubsection{Attention mechanism}
%\\paragraph{Attention mechanism.}
Attention mechanism \cite{sharma2015action} in action recognition helps to suppress irrelevant global information and highlight informative regions. Inspired by action recognition methods, iCAN \cite{gao2018ican:} exploited an instance-centric attention mechanism to enhance the information from regions and facilitate HOI classification. Furthermore, Contextual Attention \cite{wang2019deep} proposed a deep contextual attention framework for HOI detection, in which context-aware appearance features for human and object were captured. PMFNet \cite{wan2019pose} focused on pose-aware attention for HOI detection by employing human parts. Overall, methods employing attention mechanisms learn informative regions but treat each visual target (i.e., scene, human and object) separately, which are still insufficient to exploit interactive semantics for inferring HOIs.

%\subsubsection{Graph-based reasoning}
%\\paragraph{Graph-based reasoning.}
Graph Parsing Neural Network (GPNN) \cite{qi2018learning} introduced a learnable graph-based structure, in which HOIs were represented with a graph structure and parsed in an end-to-end manner. The above structure was a generalization of Message Passing Neural Network \cite{gilmer2017neural} using a message function and a Gated Recurrent Unit (GRU) to iteratively update states. GPNN was innovative but showed some limitations. Firstly, it reasoned interactive features at the coarse instance-level (i.e., each instance was encoded as an infrangible node), which suffered from handling complex interactions. In addition, it required iteratively message passing and updating. Thirdly, it excluded semantic information from the scene in inferring HOIs. Lately, RPNN \cite{zhou2019relation} introduced a complicated structure with two graph-based modules incorporated together to infer HOIs, but in which fine-grained human pose was required as prior information.

In this paper, we aim to develop a novel graph-based model to provide interactive semantic reasoning between visual targets. Instead of coarse and iterative message passing between instances, our model captures pixel-level interactive semantics between targets all at once. Furthermore, our model is free from costly annotations like human pose.

\section{Proposed Method}
\subsection{Overview of in-Graph}

The detailed design of proposed in-Graph model is provided in Figure~\ref{FIG:2}. Scene, human and object are three semantic elements been considered as three visual targets in our model, referred to as \emph{$T_{scene}$}, \emph{$T_{human}$}, and \emph{$T_{object}$}, respectively. The proposed in-Graph takes two targets once to conduct pixel-level interactive reasoning. Each of the targets takes convolutional feature \emph{$X\in\Re^{H*W*D}$} according to the corresponding boxes (i.e., the whole image, human candidate boxes and object candidate boxes) as input. Here \emph{H*W} denotes locations and \emph{D} denotes feature dimension.

We first propose a project function to map two feature tensors into a graph-based semantic space named interactive space, where a fully-connected graph structure can be built. Based on the graph structure, message passing process is then adopted as modeling the interaction among all nodes by propagating and aggregating interactive semantics. Finally, the update function provides a reversed projection over interactive space and output feature \emph{Y}, enabling us to utilize the reasoned semantics in convolution space. We then describe its architecture in details and explain how we apply it into HOI detection task.

\subsection{Project Function}

Project function aims to provide an pattern \emph{$P(·)$} to fuse two targets together, after which message passing process can be efficiently computed. The calculation process of \emph{$P(·)$}  can be divided into three parts: a feature conversion denoting as \emph{$\varphi(X_1,X_2;W_\varphi)$}, a weights inference denoting as \emph{$\theta(X_1,X_2;W_\theta)$} and a linear combination, where \emph{$W_\varphi$} and \emph{$W_\theta$} are learnable parameters. Finally, the function outputs a matrix \emph{$V=P(X_1,X_2)\in\Re^{2N*2C}$}, where \emph{$X_1$} and \emph{$X_2$} are input feature tensors, \emph{$2N$} denotes the number of nodes in interactive space and \emph{$2C$} refers to dimension.

%\subsubsection{Feature conversion}
%\paragraph{Feature conversion.}
Feature conversion. Given feature tensors of two targets \emph{$X_1,X_2\in\Re^{H*W*D}$}, we first employ \emph{$1*1$} convolutions to reduce the dimensions of \emph{$X_1$} and \emph{$X_2$} to \emph{$C$}, thus the computation of the block can be valid decreased. The obtained tensors are then reshaped from \emph{$H*W*C$} to planar \emph{$L*C$}, obtaining \emph{$X_1^r,X_2^r\in\Re^{L*C}$},where its two-dimensional location pixels of \emph{$H*W$} are converted to one-dimensional vector \emph{$L$}. After that, a concatenation operation is adopted to integrate \emph{$X_1^r$} and \emph{$X_2^r$} by dimensional \emph{$C$}, obtaining \emph{$X^r=[X_1^r,X_2^r]\in\Re^{L*2C}$}.

%\subsubsection{Weights inference}
%\paragraph{Weights inference.}
Weights inference. Here, we infer learnable projection weights \emph{$B$} so that semantic information from original features can be weighted aggregated. Instead of designing complicated calculations, we simply use convolution layers to generate the dynamically weights. In this step, \emph{$X_1$} and \emph{$X_2$} are feed into \emph{$1*1$} convolutions to obtain weight tensors with channel of N. Obtained feature tensors are then reshaped as planar \emph{$B_1  ,B_2\in\Re^{N*L}$}. Finally, integrated projection weights \emph{$B=[B_1,B_2]=[b_1,b_2,…b_i,…,b_{2N}]\in\Re^{2N*L}$} are obtained by a concatenation operation.

%\subsubsection{Linear combination}
%\paragraph{Linear combination.}
Linear combination. Since the project function involves two targets, linear combination is a necessary step to aggregate the semantic information and transform targets to the unified interactive space. In particular, node \emph{$v_i\in{V}$} in interactive space is calculated as follow. Here \emph{$x_j^r\in\Re^{1*2C}$}, \emph{$v_i\in\Re^{1*2C}$}.
\begin{equation}
v_i=\sum_{\forall j}b_{ij}x_j^r.
\label{equ1}
\end{equation}

The proposed project function is simple and fast since all parameters are end-to-end learnable and come from 1*1 convolutions. Such a function achieves semantic fusion between two targets and maps them into an interactive space effectively.

\subsection{Message Passing and Update Function}

\subsubsection{Message Passing} After projecting targets from convolution space to interactive space, we have a structured representation of a fully-connected graph \emph{$G=(V,E,A)$}, where each node contains a feature tensor as its state and all nodes are considered as fully-connected with each other. Based on the graph structure, message passing process is adopted to broadcast and integrate semantic information from all nodes over the graph.

GPNN \cite{qi2018learning} applies an iterative process with GRU to enable nodes to communicate with each other, whereas it needs to run several times iteratively towards convergence. We reason interactive semantics over the graph structure by adopting a single-layer convolution to efficiently build communication among nodes. In our model, the message passing functions\emph{$M(·)$} is computed by:
\begin{equation}
M(V)=\omega(AV)=Conv1D(V)\oplus{V}.
\label{equ2}
\end{equation}
Here \emph{$A$} denotes the adjacency matrix among nodes learned by gradient decent during training, reflecting the weights for edge \emph{$E$}. \emph{$\omega(·)$} denotes the state update of nodes. In our implementation, the operation of \emph{$Conv1D(V)$} is a channel-wise 1D convolution layer that performs Laplacian smoothing \cite{li2018deeper} and propagates semantic information among all connected nodes. After information diffusion, the \emph{$\oplus$} implements addition point to point which updates the hidden node states according to the incoming messages.

\subsubsection{Update Function} To apply above reasoning results into convolutional network, an update function \emph{$U(·)$} provides a reverse projection for reasoned nodes from interactive space to convolution space, which output \emph{$Y$} as a new feature tensor. Given the reasoned nodes \emph{$V'\in\Re^{2N*2C}$}, update function first adopts a linear combination as follows:
\begin{equation}
y_i=\sum_{\forall j}b_{ij}v'_j.
\label{equ3}
\end{equation}
Where the projection weights \emph{$B$} is transposed and reused, here \emph{$v'_j\in\Re^{1*2C}$}, \emph{$y_i\in\Re^{1*2C}$}.

After the linear combination, we reshape the obtained tensor from planar \emph{$L*2C$} to three-dimensional \emph{$H*W*2C$}. Finally, a \emph{$1*1$} convolution is attached to expand the feature dimensions from \emph{$C$} to \emph{$D$} to match the inputs. In this way, updated features in convolution space can play its due role in the following schedule.

\subsection{in-GraphNet}

\subsubsection{Assembling in-Graph Model} Our in-Graph model improves the ability of modelling HOIs by employing interactive semantic reasoning beyond stack of convolutions. It is noted that the human visual system is able to progressively capture interactive semantics from the scene and related instances to recognize a HOI. Taking the HOI triplet $\langle$ human, surf, surfboard $\rangle$ as an example, the scene-wide interactive semantics connected with the scene (e.g., sea) and instances (e.g., human, surfboard) can be captured as prior knowledge and instance-wide interactive semantics between the person and surfboard are learned to further recognize the verb (i.e., surf) and disambiguate other candidates (e.g., carry). Inspired by this human perception, we assign in-Graphs in two levels to build in-GraphNet, which are scene-wide level and instance-wide level. The scene-wide in-Graphs contain a human-scene in-Graph and an object-scene in-Graph, the instance-wide in-Graph refers to human-object in-Graph.

\begin{figure}
	\centering
		\includegraphics[scale=.23]{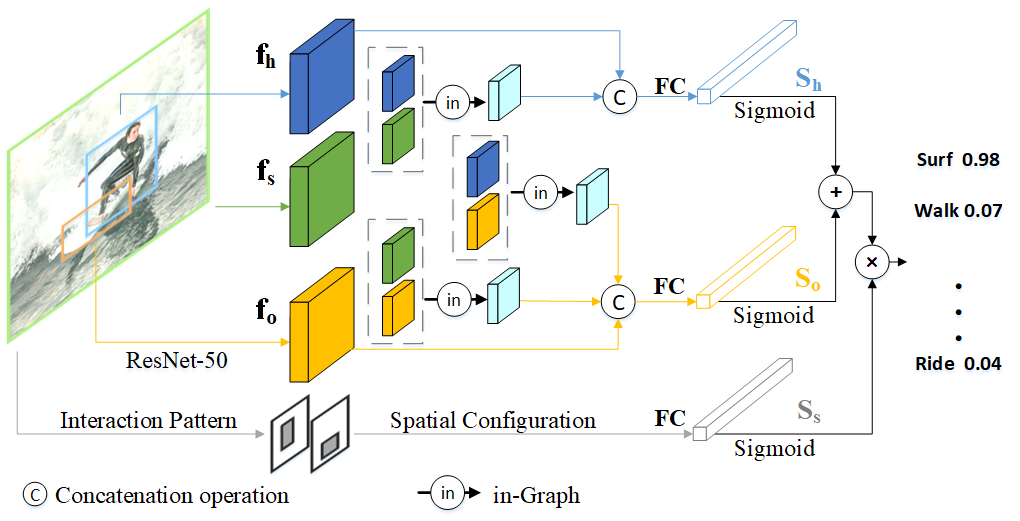}
	\caption{An overview of in-GraphNet. The framework consists of three branches, which are a human-centric branch, an object-centric branch and a spatial configuration branch. We integrate two scene-wide in-Graphs and an instance-wide in-Graph into in-GraphNet to infer HOIs.}
	\label{FIG:3}
\end{figure}

An overview of proposed in-GraphNet is shown in Figure~\ref{FIG:3}. Since in-Graph models are light-weight, it can be easily incorporated into existing CNN architectures. ResNet-50 \cite{he2016deep} is employed as the backbone network in our implementation. We denote the candidate bounding boxes of the human and object by \emph{$b_h$} and \emph{$b_o$}, respectively. We also express the box of the whole scene as \emph{$b_s$}. After the process of shared ResNet-50 C1-C4 and RoI pooling according to respective candidate boxes, ResNet-50 C5 generates input features for targets \emph{$T_{scene}$}, \emph{$T_{human}$}, and \emph{$T_{object}$}, teamed as \emph{$f_s$}, \emph{$f_h$} and \emph{$f_o$}. These three targets are then assigned into three in-Graph models in pairs. While input features for in-Graph model have dimension \emph{$D=2048$}, we set reduced dimension \emph{$C=1024$} , number of nodes \emph{$N=512$}.

\subsubsection{Three Branches} The in-GraphNet contains a human-centric branch, an object-centric branch and a spatial configuration branch, where \emph{$f_h$}, \emph{$f_o$} and the spatial configuration \cite{gao2018ican:} based on two-channel interaction pattern \cite{chao2018learning} are adopted as basic feature representations, respectively. In scene-wide interactive reasoning, semantic features obtained from human-scene in-Graph and object-scene in-Graph are concatenated to \emph{$f_h$} and \emph{$f_o$} respectively to enrich presentations in human-centric and object-centric branches. In instance-wide interactive reasoning, semantic feature output from human-object in-Graph is concatenated into the object-centric branch only, because appearance of an object is usually constant in different interactions and provides minor effect in human-centric representation. Finally, the enhanced features from three branches are fed into fully connected layers and perform classification operations. In this way, the entire framework is implemented to be fully differentiable and end-to-end trainable using gradient-based optimization. With the formulation above, rich interactive relations between visual targets can be explicitly utilized to infer HOIs.

\begin{table}
\centering
\resizebox{85mm}{!}{
\begin{tabular}{llc}
\toprule
Method & Backbone Network & mAP$_{role}$(\%)  \\ \hline
Gupta et al. \cite{yatskar2016situation}  & ResNet-50-FPN & 31.8  \\
InteractNet \cite{gkioxari2018detecting}  & ResNet-50-FPN & 40.0  \\
BAR-CNN \cite{kolesnikov2018detecting} & Inception-ResNet & 41.1  \\
GPNN \cite{qi2018learning}  & Deformable CNN & 44.0  \\
iCAN \cite{gao2018ican:}   & ResNet-50 & 45.3  \\
Contextual Att \cite{wang2019deep}   & ResNet-50 & 47.3 \\ \hline
iHOI \cite{xu2018interact}  & ResNet-50 & 40.4  \\
RPNN \cite{zhou2019relation}   & ResNet-50 & 47.5  \\
TIN(R$P_dC_d$) \cite{li2019transferable}   & ResNet-50 & \underline{47.8}  \\ \hline
our baseline   & ResNet-50 & 44.8  \\
\textbf{In-GraphNet (ours)} & ResNet-50 & \textbf{48.9}  \\
\bottomrule
\end{tabular}}
\caption{Performance comparison with state-of-the-arts on V-COCO dataset.}
\label{table1}
\end{table}

\begin{table}
\centering
\resizebox{85mm}{!}{
\begin{tabular}{llccc}
\toprule
\multirow{2}{*}{Method} & \multirow{2}{*}{Backbone Network} & \multicolumn{3}{c}{Default} \\ \cline{3-5}
 & & \multicolumn{1}{c}{full} & \multicolumn{1}{c}{rare} & \multicolumn{1}{c}{non-rare} \\ \hline
Shen et al. \cite{shen2018scaling}  & VGG-19 &6.46 &4.24 &7.12 \\
HO-RCNN \cite{chao2018learning}  & CaffeNet & 7.81 & 5.37 & 8.54  \\
InteractNet \cite{gkioxari2018detecting}  &ResNet-50-FPN &9.94 &7.16 &10.77 \\
GPNN \cite{qi2018learning}  &Deformable CNN & 13.11 & 9.34 & 14.23 \\
iCAN \cite{gao2018ican:}  & ResNet-50 & 14.84 & 10.45 & 16.15  \\
Contextual Att \cite{wang2019deep}  & ResNet-50 & 16.24 & 11.16 & 17.75\\ \hline
iHOI \cite{xu2018interact}  &ResNet-50 & 9.97 & 7.11  &10.83  \\
RPNN \cite{zhou2019relation}  &ResNet-50 & \underline{17.35} & 12.78  & \underline{18.71} \\
TIN(R$P_dC_d$) \cite{li2019transferable}  &ResNet-50 & 17.03 & \textbf{13.42}  \\ \hline
our baseline  &ResNet-50 & 15.41 & 10.71  & 16.81\\
\textbf{In-GraphNet (ours)}  & ResNet-50 & \textbf{17.72} & \underline{12.93} & \textbf{19.31}  \\
\bottomrule
\end{tabular}}
\caption{Results on HICO-DET test set. Default: all images. Full: all 600 HOI categories. Rare: 138 HOI categories with less than 10 training instances. Non-Rare: 462 HOI categories with 10 or more training instances.}
\label{table2}
\end{table}

\subsubsection{Training} Since HOI detection is a multi-label classification problem where more than one HOI label might be assigned to a $\langle$ human, object $\rangle$ candidate, our model is trained in a supervised fashion using the multi-label binary cross-entropy loss. All three branches are trained jointly, where the overall loss for each interaction category is the sum of three losses from three branches.

For each image, pairwise candidate boxes (\emph{$B_h*B_o$}) for each HOI category are assigned binary labels based on the prediction. Similar to general HOI detection frameworks \cite{gkioxari2018detecting,gao2018ican:}, we use a fusion of scores output from each branch to predict a final score for each HOI.
\begin{equation}
S_{h,o}=S_h^{pre}*S_o^{pre}*(S_h+S_o)*S_s.
\label{equ3}
\end{equation}
Here \emph{$S_h$}, \emph{$S_o$} and \emph{$S_s$} denotes scores output from binary sigmoid classifiers in human-centric branch, object-centric branch and spatial configuration branch, respectively. \emph{$S_h^{pre}$} and \emph{$S_o^{pre}$} are object detection scores output from object detector for candidate boxes. \emph{$S_{h,o}$} is the final score for each HOI.

\section{Experiments and Evaluations}
\subsection{Experimental Setup}

\begin{figure}
	\centering
		\includegraphics[scale=.24]{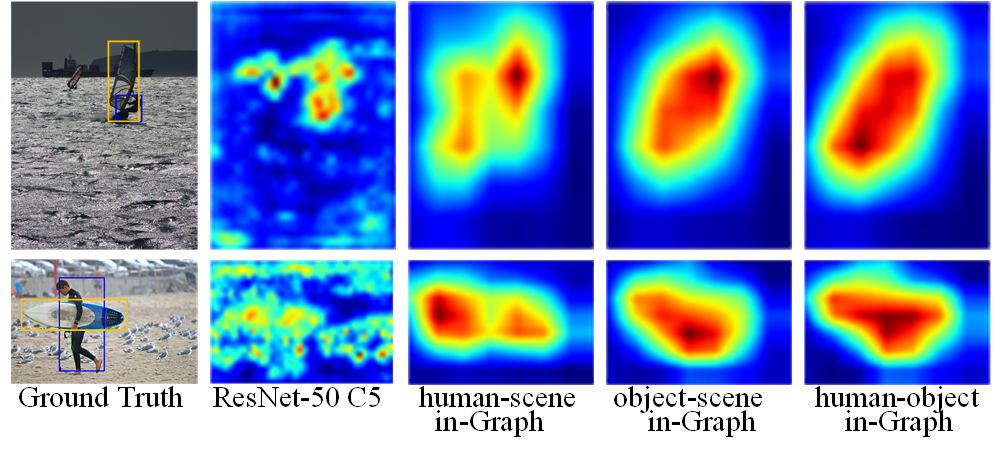}
	\caption{in-Graph visualization. The first column shows two HOI examples (i.e., $\langle$ human, surf, surfboard $\rangle$, $\langle$ human, hold, surfboard $\rangle$ ). The rest of columns visualize feature responses from ResNet-50 C5, human-scene in-Graph, object-scene in-Graph and human-object in-Graph, respectively.}
	\label{FIG:4}
\end{figure}

%\subsubsection{Datasets and evaluation metrics}
%\paragraph{Datasets and evaluation metrics.}
Datasets and evaluation metrics. We evaluate our model and compare it with the state-of-the-arts on two large-scale benchmarks, including V-COCO \cite{yatskar2016situation} and HICO-DET \cite{chao2018learning} datasets. V-COCO \cite{yatskar2016situation} includes 10,346 images, which is a subset of MS COCO dataset. It contains 16,199 human instances in total and provides 26 common HOI annotations. HICO-DET \cite{chao2018learning} contains about 48k images and 600 HOI categories over 80 object categories, which provides more than 150K annotated $\langle$ human, object $\rangle$ pairs. We use role mean average precision (role mAP) \cite{yatskar2016situation} on both benchmarks.

%\subsubsection{Implementation details}
%\paragraph{Implementation details.}
Implementation details. Following the protocol in \cite{gao2018ican:}, human and object bounding boxes are generated using the ResNet-50 version of Faster R-CNN \cite{ren2017faster}. Human boxes with scores higher than 0.8 and object boxes with scores higher than 0.4 are kept for detecting HOIs. We train our model with Stochastic Gradient Descent (SGD), using a learning rate of 1e-4, a weight decay of 1e-4, and a momentum of 0.9. The strategy of interactiveness knowledge training \cite{li2019transferable} is adopted in our training and the model is trained for 300K and 1800K iterations on V-COCO and HICO-DET, respectively. All our experiments are conducted by tensorflow on a GPU of GeForce GTX TITAN X.

\begin{figure*}
	\centering
		\includegraphics[scale=.35]{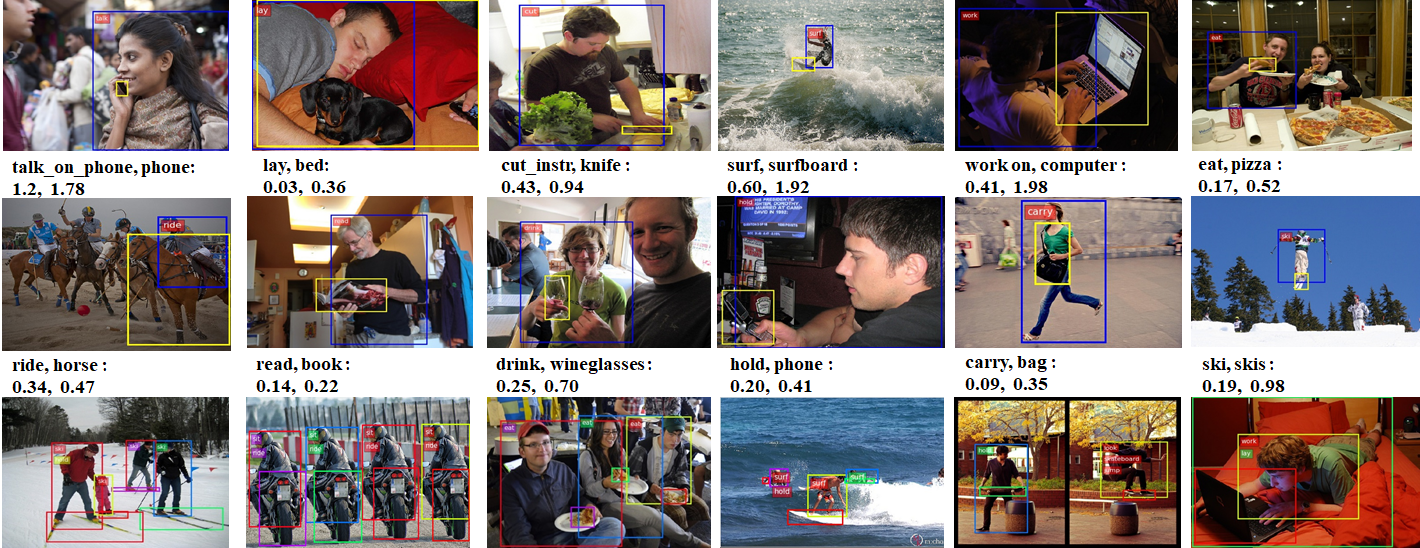}
	\caption{Visualization of HOI detections. The first and second rows show results compared with baseline on V-COCO. Each subplot displays one detected $\langle$ human, verb, object $\rangle$ triplet for easy observation. Texts below indicate the $\langle$ verb, object $\rangle$ tuple and two numbers in turn represent scores predicted by baseline and our approach. The last row shows multiple people take interactions with various objects concurrently detected by our method.}
	\label{FIG:5}
\end{figure*}

\subsection{Overall Performance}

We compare our method with several state-of-the-arts in this subsection. Methods being compared are classified into two categories, i.e., methods that are free from human pose (top of tables) and methods relying on additional pose estimators (middle of tables). Meanwhile, we strip all modules related to in-Graphs from our proposed framework as the baseline. Comparison results on V-COCO and HICO-DET in terms of \emph{$mAP_{role}$} are shown in Table~\ref{table1} and Table~\ref{table2}, respectively.

Firstly, our in-GraphNet obtains 48.9 mAP on V-COCO and 17.72 mAP on HICO-DET (Defualt full mode) and achieves absolute gains of 4.1 points and 2.3 points compared with the baseline, which are relative improvements of 9.4\% and 15\%. Besides, although pose-based methods usually perform better than pose-free methods, our method as a pose-free one outperforms all the others with the best performance, validating its efficacy for HOI detection task.

\subsection{Ablation Studies}

\begin{table}
\centering
\resizebox{85mm}{!}{
\begin{tabular}{lcccc}
\toprule
number of nodes (N) & 128 & 256 & 512 & 1024  \\ \hline
mAP$_{role}$(\%)  & 47.7 &48.2 & 48.9 & 48.7 \\
\bottomrule
\end{tabular}}
\caption{Building interactive spaces with different number of nodes.}
\label{table3}
\end{table}

We adopt several ablation studies in this subsection. VCOCO serves as the primary testbeds on which we further analyze the individual effect of components in our method.

%\subsubsection{Comprehending in-Graph}
%\paragraph{Comprehending in-Graph.}
Comprehending in-Graph. We show two examples in Figure~\ref{FIG:4} to visualize the effects of three in-Graphs been adopted. The brightness of the pixel indicates how much the feature been noticed. Intuitively, the three in-Graphs learn different interactive semantics from pairwise reasoning. Human-scene in-Graph and Object-scene in-Graph exploit interactive semantics between scene and instances. Human-object in-Graph, on the other hand, mostly focuses on the regions roughly correspond to the on-going action between human and object. In addition, to further dissect in-Graph model, empirical tests have been conducted to study the effect of number of nodes in interactive spaces. As summarized in Table~\ref{table3}, we get the best result when \emph{$N$} is set as the value of 512.

%\subsubsection{Effects of adopting different in-Graph models}
%\paragraph{Effects of adopting different in-Graph models.}
Effects of adopting different in-Graph models. As shown in Table~\ref{table4}, while directly concatenating \emph{$[f_h,f_s]$} in human-centric branch and  \emph{$[f_o,f_s, f_h]$} in object-centric branch, we can see that simply concatenating different targets without interactive reasoning barely improves the result. When we only adopt the scene-wide in-Graphs or the instance-wide in-Graph, the mAP are 48.3 and 47.7 respectively, indicating the respective effects of these two parts. Specifically, we can draw a relatively conclusion from detailed class-wise results that scene-wide in-Graphs are more adept in modelling interactions closely related to the environment, while instance-wide in-Graph performs better in depicting interactions closely related to human pose.

\begin{table}
\centering
\resizebox{85mm}{!}{
\begin{tabular}{lccccc}
\toprule
feature concatenation  & & {$\surd$} &  & &    \\
scene-wide in-Graphs   & &  & {$\surd$} & & {$\surd$}    \\
instance-wide in-Graph  & & &  & {$\surd$} & {$\surd$}    \\   \hline
mAP$_{role}$(\%)     & 44.8 & 45.2  & 48.3 & 47.7 & 48.9     \\
\bottomrule
\end{tabular}}
\caption{Impact of adopting different in-Graph models.}
\label{table4}
\end{table}

\subsection{Quantitive Examples}

For visualization, several examples of detection are given in Figure~\ref{FIG:5}. We first compare our results with baseline to demonstrate our improvements. We can see from the first two rows that our method is capable of detecting various HOIs with higher scores. In addition, the third row shows that our method can adapt to complex environments to detect multiple people taking different interactions with diversified objects.

\section{Conclusion}

In this paper, we propose a graph-based model to address the problem of lack interactive reasoning in existing HOI detection methods. Beyond convolutions, our proposed in-Graph model efficiently reasons interactive semantics among visual targets by three procedures, i.e., a project function, a message passing process and an update function. We further construct an in-GraphNet assembling two-level in-Graph models in a multi-stream framework to parse scene-wide interactive semantics and instance-wide interactive semantics for inferring HOIs. The in-GraphNet is free from costly human pose and end-to-end trainable. Extensive experiments have been conducted to evaluate our method on two public benchmarks, including V-COCO and HICO-DET. Our method outperforms both existing human pose-free and human pose-based methods, validating its efficacy in detecting HOIs.

% {\em submissions} % {\bf anything}

\section*{Acknowledgments}This paper was partially supported by National Engineering Laboratory for Video Technology - Shenzhen Division, and Shenzhen Municipal Development and Reform Commission (Disciplinary Development Program for Data Science and Intelligent Computing). Special acknowledgements are given to AOTO-PKUSZ Joint Lab for its support.

%%%The preparation of these instructions and the Francisco Cruz-Mencia.

\bibliographystyle{named}
\bibliography{inGraphReference}

\end{document}